\documentclass[a4paper, 10pt, conference]{ieeeconf}      

\IEEEoverridecommandlockouts                              

\usepackage{graphics} 
\usepackage{epsfig} 
\usepackage{mathptmx} 
\usepackage{times} 
\usepackage{amsmath} 
\usepackage{amssymb}  
\usepackage{cite}
\usepackage{wasysym} 
\usepackage{hyperref}
\usepackage{tikz}

\usepackage{cite}

\usepackage[top=1.46in, bottom=0.79in,left=0.77in, right=0.52in ]{geometry}
    
\makeatletter
\def\@citex[#1]#2{\leavevmode
\let\@citea\@empty
\@cite{\@for\@citeb:=#2\do
{\@citea\def\@citea{,\penalty\@m\ }%
\edef\@citeb{\expandafter\@firstofone\@citeb\@empty}%
\if@filesw\immediate\write\@auxout{\string\citation{\@citeb}}\fi
\@ifundefined{b@\@citeb}{\hbox{\reset@font\bfseries ?}%
\G@refundefinedtrue
\@latex@warning
{Citation `\@citeb' on page \thepage \space undefined}}%
{\@cite@ofmt{\csname b@\@citeb\endcsname}}}}{#1}}
\makeatother

\newcommand\Tstrut{\rule{0pt}{2.0ex}}         
\newcommand\Bstrut{\rule[-0.9ex]{0pt}{0pt}} 
\newcommand\Mstrut{\rule[-0.0ex]{0pt}{0pt}}

\newcommand{\fig}[1]{\mbox{Fig. \ref{#1}}}
\newcommand{\tab}[1]{\mbox{Table \ref{#1}}}

\newcommand\copyrighttext{%
  \footnotesize This work has been submitted to the IEEE for possible publication. Copyright may be transferred without notice, after which this version may no longer be accessible.}%
\newcommand\copyrightnotice{%
\begin{tikzpicture}[remember picture,overlay]%
\node[anchor=south,yshift=10pt] at (current page.south) {\fbox{\parbox{\dimexpr\textwidth-2cm}{\copyrighttext}}};%
\end{tikzpicture}%
\vspace{-10pt}%
}

\title{\LARGE \bf
Robust Semantic Segmentation in Adverse Weather Conditions by means of Fast Video-Sequence Segmentation
}

\author{Andreas Pfeuffer$^{1}$ and Klaus Dietmayer$^{1}$
\thanks{$^{1}$Andreas Pfeuffer, and Klaus Dietmayer are with the Institute of Measurement, Control, and Microtechnology, Ulm University, 89081 Ulm, Germany, firstname.lastname@uni-ulm.de}%
}

\begin{document}

\maketitle
\copyrightnotice
\thispagestyle{empty}
\pagestyle{empty}

\begin{abstract}
	Computer vision tasks such as semantic segmentation perform very well in good weather conditions, but if the weather turns bad, they have problems to achieve this performance in these conditions. One possibility to obtain more robust and reliable results in adverse weather conditions is to use video-segmentation approaches instead of commonly used single-image segmentation methods. Video-segmentation approaches capture temporal information of the previous video-frames in addition to current image information, and hence, they are more robust against disturbances, especially if they occur in only a few frames of the video-sequence. 
	However, video-segmentation approaches, which are often based on recurrent neural networks, cannot be applied in real-time applications anymore, since their recurrent structures in the network are computational expensive.
	For instance, the inference time of the LSTM-ICNet, in which recurrent units are placed at proper positions in the single-segmentation approach ICNet, increases up to 61 percent compared to the basic ICNet. Hence, in this work, the LSTM-ICNet is sped up by modifying the recurrent units of the network so that it becomes real-time capable again. Experiments on different datasets and various weather conditions show that the inference time can be decreased by about 23 percent by these modifications, while they achieve similar performance than the LSTM-ICNet and outperform the single-segmentation approach enormously in adverse weather conditions.
\end{abstract}


\section{Introduction}

\begin{figure}[tbp]
		\includegraphics[width=1.0\columnwidth]{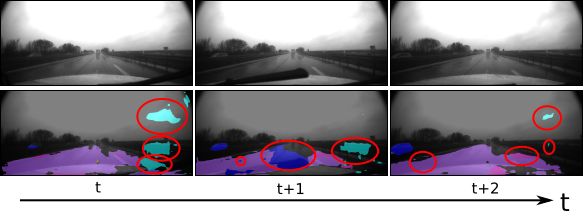}
		\caption{Semantic segmentation of a video sequence by means of the single-segmentation approach ICNet \cite{Zhao_2017_ICNet_forRealTimeSemanticSegmentationOnHighResolutionImages} in rain}
		\label{fig_introduction_segmentation}
		\vspace{-5mm}
	\end{figure}

	A robust semantic segmentation in all weather conditions is an important requirement for autonomous driving applications to yield a reliable environment perception of the car's surrounding. 
	Today's semantic segmentation approaches such as ICNet \cite{Zhao_2017_ICNet_forRealTimeSemanticSegmentationOnHighResolutionImages}, PSPNet \cite{Zhao_2017_PyramidScenParsingNetwork}, and Deeplab \cite{Chen_2018_Deeplabv3p_EncoderDecoderWithAtrousSeparableConvolutionForSemanticImageSegmentation} rely on neural networks and achieve very good results in good weather conditions. However, if the weather conditions are not optimal, e.g. in fog, rain, at night or even in blinding sun, they fail, and their performance decreases enormously. The reason is that video sequences are often segmented by processing each frame independently from each other.
	Due to this, the segmentation results are often varying between the single frames of the video sequence, especially, if one or several images are disturbed due to adverse weather effects. For instance, the object edges are flickering and ghost objects occur in some frames. Furthermore, some parts of the objects are misclassified in one frame, while they are detected correctly in the next frame. In \fig{fig_introduction_segmentation}, some of these typical errors are shown in the case of heavy rain. For example, parts of the road are misclassified, and parts of the sky are wrongly predicted as truck. In most cases, these segmentation errors occur only in one or in a few frames of the video sequence, while they are classified correctly in the next frames. These errors can be reduced by the use of previous image information in addition to the current image information. 
	One possibility to capture these temporal image information is to use recurrent neural networks (RNNs), e.g by means of convolutional LSTMs (convLSTMs, \cite{Shi_2015_ConvolutionalLSTMNetwork_AMachineLearningApproachForPrecipitationNowcasting}). For example, in \cite{Pfeuffer_2019_SemanticSegmentationOfVideoSequencesWithConvolutionalLSTMs}, it was shown that the segmentation accuracy of the ICNet \cite{Zhao_2017_ICNet_forRealTimeSemanticSegmentationOnHighResolutionImages} can be increased in good weather conditions, if convLSTM cells are placed at suitable positions in the network. 
	However, the performance increase is at the expense of the inference time, e.g. it increases up to 61 percent compared to the basic segmentation approach. Therefore, this video-segmentation approach based on RNN is not suitable any more for real-time applications such as autonomous driving due to its high computational costs.
	 
	In this work, two real-time capable extensions of the video-segmentation approach LSTM-ICNet are introduced, the Fast-LSTM-ICNet and the Faster-LSTM-ICNet. Furthermore, we show, that these video-segmentation approaches perform much better than single-image approaches, if one or several frames of the video are disturbed, e.g. due to adverse weather conditions.


\section{Related Work}

	Currently segmentation approaches perform very well in good weather conditions, as the high scores on several segmentation benchmarks show, and also can be applied to real-time applications such as mobile robots or autonomous cars with minimal performance loss. However, they have problems to achieve this performance in adverse weather conditions due to sensor disturbances and poor observability of objects caused by darkness, rain, snow or fog. Nevertheless, a good segmentation accuracy is still required in these conditions, and hence, different possibilities are analyzed in recent years on how to obtain a robust segmentation approach performing well in adverse weather conditions. 
	A common way is to simulate adverse weather effects as realistic as possible, and to train the network with this data, in hope, that the network also performs well in real adverse weather conditions. For instance, Sakaridis et al. \cite{Sakaridis_2018_SemanticFoggySceneUnderstandingWithSyntheticData} simulate fog by means of the depth values. They show that training on the Cityscapes dataset with this synthetic fog increases the performance in real fog scenarios. In \cite{Porav_2019_ICanSeeClearlyNow_ImageRestorationViaDeRaining}, a neural network is trained by means of simulated rain data to derain images, so that the following segmentation network can process these undisturbed images.	
	Another possibility is to use data of different sensors to increase the segmentation accuracy. The motivation is that different sensors perform differently in diverse weather conditions. For example, in the case of blinding sun, the camera is disturbed and white areas occur in the camera image, while the environment can be still observed by the lidar sensor without disturbances. Hence, in \cite{Pfeuffer_2019_RobustSemanticSegmentationInAdverseWeatherConditionsByMeansOfSensorDataFusion}, a segmentation approach was developed, which fuses camera and lidar data to obtain a more robust segmentation approach. It turns out that the performance increases enormously in adverse weather conditions, if one sensor is disrupted by unknown noise.
	
	In this work, we focus on a third possibility, the video-segmentation. The idea is that the segmentation errors can be compensated by the image informations of previous video frames. 
	One of the first video-segmentation approaches was proposed by Valipour et al. \cite{Valipour_2017_RecurrentFullyConvolutionalNetworksForVideoSegmentation}. The authors extended the Fully Convolutional Network (FCN) \cite{Long_2015_FullyConvolutionalNetworksForSemanticSegmentation} by a convLSTM cell \cite{Shi_2015_ConvolutionalLSTMNetwork_AMachineLearningApproachForPrecipitationNowcasting} placed between the encoder and decoder. Evaluations on several benchmarks show that the performance increases by this method. In \cite{Simonyan_2015_VeryDeepConvolutionalNetworksForLargeScaleImageRecognition}, Yurdakul et al. replaced the FCN by a modified VGG19 architecture. Moreover, they compare different recurrent units, e.g. convRNN, convGRU, and convLSTMs, and figured out, that the convLSTM-cells perform best. Furthermore, the segmentation approach developed by Nabavi et al. \cite{Nabavi_2018_FutureSemanticSegmentationWithConvolutionalLSTM}, which is able to predict the future segmentation map by means of previous image information, is based on a ResNet encoder-decoder architecture, in which  convLSTM cells are placed between each skip connection and between the encoder and decoder. 
	In 2019, the LSTM-ICNet was proposed in \cite{Pfeuffer_2019_SemanticSegmentationOfVideoSequencesWithConvolutionalLSTMs}, in which the single-segmentation approach ICNet \cite{Zhao_2017_ICNet_forRealTimeSemanticSegmentationOnHighResolutionImages} is extended by convLSTM layers at suitable positions.
	However, the inference time increases enormously compared to the original ICNet even in the case of only one added convLSTM layer, since convLSTMs are computational expensive in general. Hence, in \cite{pfeuffer_2019_SeparableConvolutionalLSTMsForFasterVideoSegmentation}, the convLSTM cells are sped up by replacing each convolution within the convLSTM cell by depthwise separable convolutions. Thus the inference time was reduced by up to 15 percent on a GPU. In this work, it is described how to reduce the inference time further by means of small network modifications of the LSTM-ICNet.


\section{Network Architecture}

	In this section, the network architecture of the proposed video-segmentation approaches Fast-LSTM-ICNet and Faster-LSTM-ICNet are described in more detail. Both approaches are accelerated versions of the LSTM-ICNet \cite{Pfeuffer_2019_SemanticSegmentationOfVideoSequencesWithConvolutionalLSTMs}, and hence, a short overview of the LSTM-ICNet is given at first.

\subsection{LSTM-ICNet}
	
\begin{figure}[tbp]
		\includegraphics[width=1.0\columnwidth]{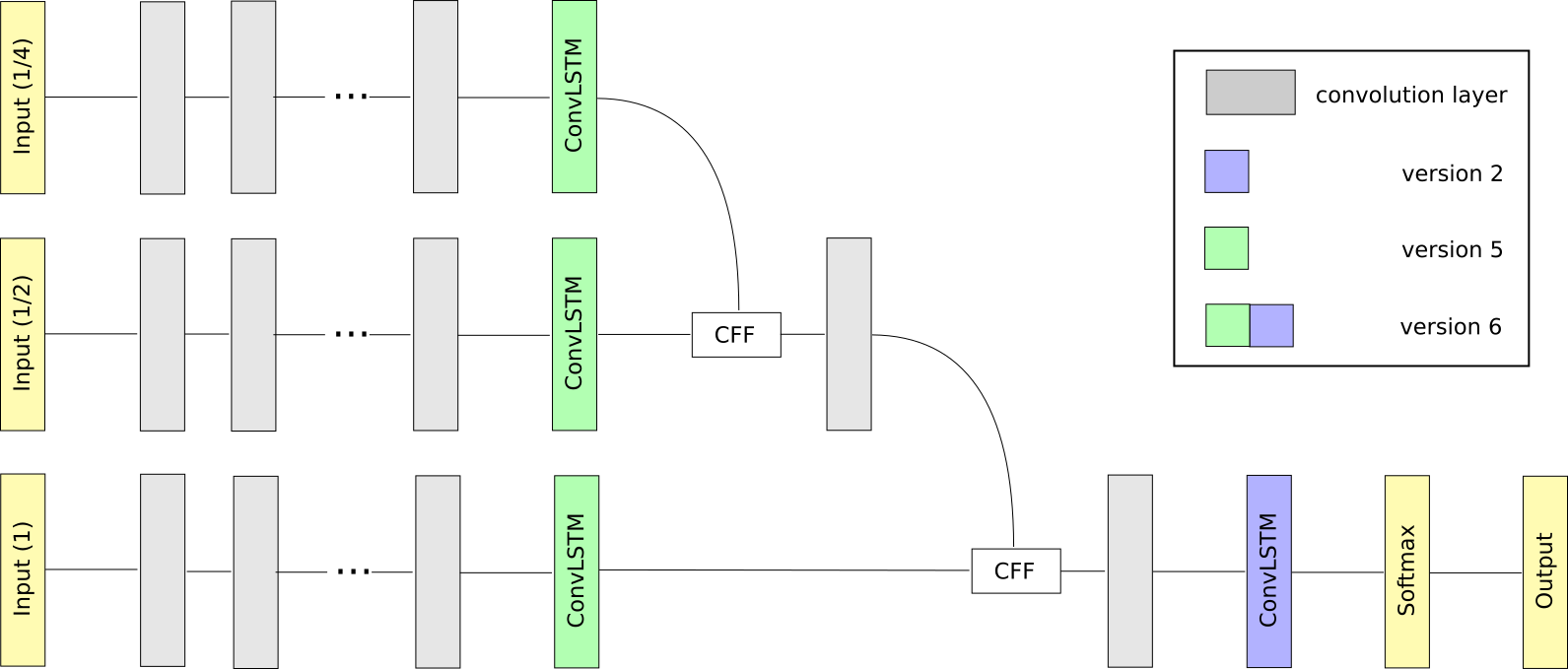}
		\caption{Network architecture of the LSTM-ICNet \cite{Pfeuffer_2019_SemanticSegmentationOfVideoSequencesWithConvolutionalLSTMs}. The original ICNet structure is illustrated by gray boxes, while the colored boxes represent different positions of the recurrent units}
		\label{fig_LSTM_ICNet_architecture}
		\vspace{-5mm}
	\end{figure}

\begin{figure*}[tbp]
		\includegraphics[width=1.0\textwidth]{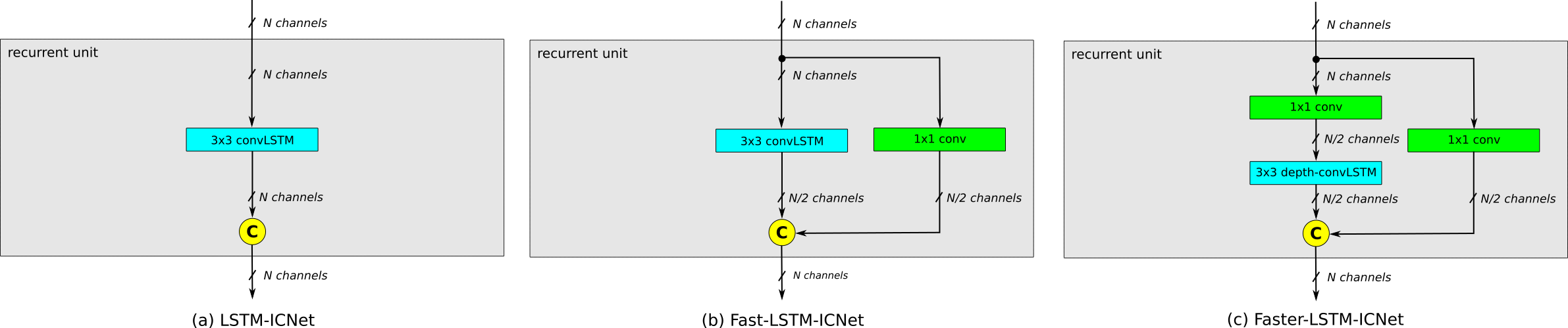}
		\caption{(a) Recurrent unit of the LSTM-ICNet \cite{Pfeuffer_2019_SemanticSegmentationOfVideoSequencesWithConvolutionalLSTMs} (b) recurrent unit of the Fast-LSTM-ICNet (c) recurrent unit of the Faster-LSTM-ICNet}
		\label{fig_FastLSTMICNet_recurrentUnit}
		\vspace{-5mm}
	\end{figure*}

	The LSTM-ICNet is based on the ICNet \cite{Zhao_2017_ICNet_forRealTimeSemanticSegmentationOnHighResolutionImages}, a real-time capable segmentation method, which processes the input image in separate network branches for  different image resolutions, and combines the determined features of each branch by means of Cascade Feature Fusion (CFF) layers. In \cite{Pfeuffer_2019_SemanticSegmentationOfVideoSequencesWithConvolutionalLSTMs}, six different LSTM-ICNet versions are described, however in this work, we focus on \mbox{version 2}, \mbox{version 5} and \mbox{version 6}, which turned out to perform best. In \mbox{version 2}, the recurrent unit is placed directly before the softmax layer at the end of the network, which corresponds to a temporal filtering of the segmentation result, and in \mbox{version 5}, a recurrent unit is located at the end of each resolution branch. Finally, \mbox{version 6} is a combination of \mbox{version 2} and \mbox{version 5}, i.e. there are recurrent units at the end of each resolution branch and directly before the softmax layer. The network architecture of the described versions is illustrated in \fig{fig_LSTM_ICNet_architecture}.  
	The recurrent units (see \fig{fig_FastLSTMICNet_recurrentUnit}(a)) consist of a simple convLSTM cell using a kernel size of $3 \times 3$. Furthermore, their amount of output channels $O$ is equal to the number of input layers $I$. According to \cite{pfeuffer_2019_SeparableConvolutionalLSTMsForFasterVideoSegmentation}, 
	\begin{align}
		\left( 16 \cdot K_x \cdot K_y \cdot I + 37 \right) \cdot O \cdot D_x \cdot D_y
	\end{align}
	FLOPs are required for this recurrent unit, where $D_x \times D_y$ is the size of the feature map and $K_x \times K_y$ the kernel size of the convLSTM-cell.

\subsection{Fast-LSTM-ICNet}

	The problem of the LSTM-ICNet is that the inference time increases enormously due to the extended convLSTM cells, since convLSTMs are computationally expensive. Hence, the Fast-LSTM-ICNet is proposed, which is computationally more efficient.
	Generally, the network architecture of the Fast-LSTM-ICNet is identical to the one of the LSTM-ICNet, but the recurrent units are sped up by means of small modifications. In more detail, the channel size of the convLSTM layer is halved to save computation time. Moreover, a $1 \times 1$ convolution is applied in parallel, and the corresponding output is concatenated with the $N/2$-dimensional output of the convLSTM-cell, so that the output channel size of the recurrent unit is equal to the number of the input channel size similar to the LSTM-ICNet. The architecture of the described recurrent recurrent unit is illustrated in \fig{fig_FastLSTMICNet_recurrentUnit}(b).
	By this modification, only
	\begin{align}
		\left( \left( 16 \cdot K_x \cdot K_y \cdot I + 37 \right) \cdot \frac{O}{2} + 2 \cdot I \cdot \frac{O}{2} \right) \cdot D_x \cdot D_y
	\end{align}
	FLOPs have to be applied for the recurrent unit. This corresponds to a theoretical speed-up of about $50\%$ in case of $I = O = 128$ and $K_x = K_y = 3$.

\subsection{Faster-LSTM-ICNet}

	The Faster-LSTM-ICNet is a further development of the Fast-LSTM-ICNet. Its idea is to speed up the recurrent units further by replacing the convLSTM layer of the recurrent unit by a depthwise separable convLSTM layer (depth-convLSTM, \cite{pfeuffer_2019_SeparableConvolutionalLSTMsForFasterVideoSegmentation}), since depth-convLSTMs are computationally more efficient than standard convLSTMs. 
	The problem is that the depth-convLSTM cell requires the same number of input and output channels analogously to the depthwise convolution. Hence, a further $1 \times 1$ convolution is added before the depth-convLSTM, which reduces the input channel size of the recurrent unit to an appropriate input-size for the depth-convLSTM cell. The output of the convLSTM cell is concatenated with the $1 \times 1$ convolution of the parallel branch.
	An overview of the described recurrent unit architecture is shown in \fig{fig_FastLSTMICNet_recurrentUnit}(c). All in all, this recurrent unit takes about 
	\begin{align}
		\left( \left( 2 \cdot I + 16 \cdot K_x \cdot K_y + 37 \right) \cdot \frac{O}{2} + 2 \cdot I \cdot \frac{O}{2} \right) \cdot D_x \cdot D_y
	\end{align}
	FLOPs, and the computational costs amount to only $1.88\%$ of the recurrent unit of the LSTM-ICNet, and $3.70\%$ of the recurrent unit of the Fast-LSTM-ICNet, if $I = O = 128$ and $K_x = K_y = 3$.


\section{Evaluation}

	 The proposed video-segmentation approaches are now evaluated qualitatively and quantitatively on two different datasets, the Cityscapes dataset \cite{cityscape_dataset} and the AtUlm-Dataset \cite{Pfeuffer_2019_RobustSemanticSegmentationInAdverseWeatherConditionsByMeansOfSensorDataFusion}. Both datasets contain fine-annotated camera images and the corresponding images of the previous time steps. The Cityscapes dataset was recorded in several German cities and consists of about 3000 images of size $1024 \times 2048$ for training. The proposed approaches are then evaluated on the 500 validation images. Note, that only 19 of the provided 30 classes are used analogously to \cite{Pfeuffer_2019_RobustSemanticSegmentationInAdverseWeatherConditionsByMeansOfSensorDataFusion, Pfeuffer_2019_SemanticSegmentationOfVideoSequencesWithConvolutionalLSTMs, Zhao_2017_ICNet_forRealTimeSemanticSegmentationOnHighResolutionImages}. 
	 The AtUlm-Dataset contains 2011 images of size $512 \times 1920$ recorded in optimal weather conditions in the surrounding of Ulm (Germany), where 1502 images are used for training and 509 images for validation. Furthermore, there are 398 images recorded in adverse weather conditions such as rain, darkness and fog, which are used to evaluate the robustness of the proposed approaches in real adverse weather conditions. The training and validation set consist of diverse sequences recorded at several different locations and the frame-rate of the camera concerns $15Hz$.
	 
	 The proposed approaches are evaluated by means of the popular evaluation metrics pixelwise accuracy and mean Intersection over Union (mIoU). 
	 In this work, small video-sequences of length four are used, but similar to \cite{Pfeuffer_2019_SemanticSegmentationOfVideoSequencesWithConvolutionalLSTMs}, only the fourth frame is considered for the training loss. 
	 Analogously to several image processing methods, the video sequences are randomly flipped and scaled during training to avoid overfitting.
	 Additionally, an advanced data augmentation method is used similar to \cite{Pfeuffer_2019_RobustSemanticSegmentationInAdverseWeatherConditionsByMeansOfSensorDataFusion} to  increase the robustness of the segmentation approach. 
	 In more detail, different disturbances are applied at random to one frame, to several frames, or to every frame of the video-sequence. For example, noise such as Gaussian noise and salt-and-pepper noise, is added to the camera images. Furthermore, white polygons of random size are fitted to the video frames, and the image brightness is reduced randomly.  
	 Generally, the hyper-parameters are similarly chosen as in our previous work \cite{Pfeuffer_2019_SemanticSegmentationOfVideoSequencesWithConvolutionalLSTMs}, but some training parameters are improved. For instance, the Stochastic Gradient Descent is replaced by the Adam Optimizer, since the Adam Optimizer is more suitable for recurrent neural networks, and gradient clipping is applied during training. 
	 Furthermore, the initial learning rate is set to $1e-5$, and is reduced according to the poly-learning rate policy for the $60k$ training iterations. Due to memory reasons, a batch size of one was used for the Cityscapes dataset and a batch size of two for the AtUlm-Dataset, and hence, the performance is not as good as known from the popular datasets. For example, the ICNet achieves a mIoU-value of $67.7\%$ using a batch size of 16, but the performance decreases by about 8 percent to $59.9\%$ using a batch size of 1. The proposed video-segmentation approaches are implemented in Tensorflow \cite{tensorflow}, and the corresponding source code is publicly available on \url{https://github.com/Andreas-Pfeuffer/LSTM-ICNet}.
	 Moreover, the network parameters are initialized with the same  pretrained model, but the parameters of the recurrent unit are initialized randomly. Each training is repeated for five times to compensate the training fluctuations, and thus, the average value and the standard deviation of the training runs are given hereinafter.
	 
	 In the following, the proposed Fast-LSTM-ICNet and the Faster-LSTM-ICNet versions are evaluated and compared with the original ICNet and LSTM-ICNet. Furthermore, they are evaluated in different use-cases, e.g in good and adverse weather conditions, and if one or several frames of the video-sequence are disturbed by unknown noise. In addition, the inference time of the proposed segmentation approaches is discussed.

\subsection{Inference Time Analysis}

	An important property of segmentation approaches in general is its inference time and if they can be applied to real-time applications. Hence, \tab{table_results_Cityscapes} contains the inference time of all considered approaches using an image size of $1024 \times 2048$, which were applied on a single Nvidia Titan X. Generally, the LSTM-ICNet versions are more computational expensive than the single-segmentation approach ICNet, since the computation time increases up to $61 \%$, which corresponds to an increase of $27ms$. 
	Furthermore, the time analysis indicates that the Fast-LSTM-ICNet and Faster-LSTM-ICNet are computationally more efficient than the corresponding LSTM-ICNet versions. For instance, the inference time is reduced by $12 \%$ from $72.1 ms$ to $63.5 ms$ in the case of Fast-LSTM-ICNet \mbox{version 6} and by about $23 \%$ from $72.1 ms$ to $55.8 ms$ in the case of Faster-LSTM-ICNet \mbox{version 6}. The most time-efficient video-segmentation approach is Faster-LSTM-ICNet \mbox{version 2}, which inference time takes only about $7 ms$ longer than the ICNet and performs only slightly worse than the corresponding LSTM-ICNet \mbox{version 2}, as the results in \tab{table_results_Cityscapes} show.

\subsection{Evaluation in Simulated Adverse Weather Conditions}

\begin{figure}[tp]
		\includegraphics[width=1.0\columnwidth]{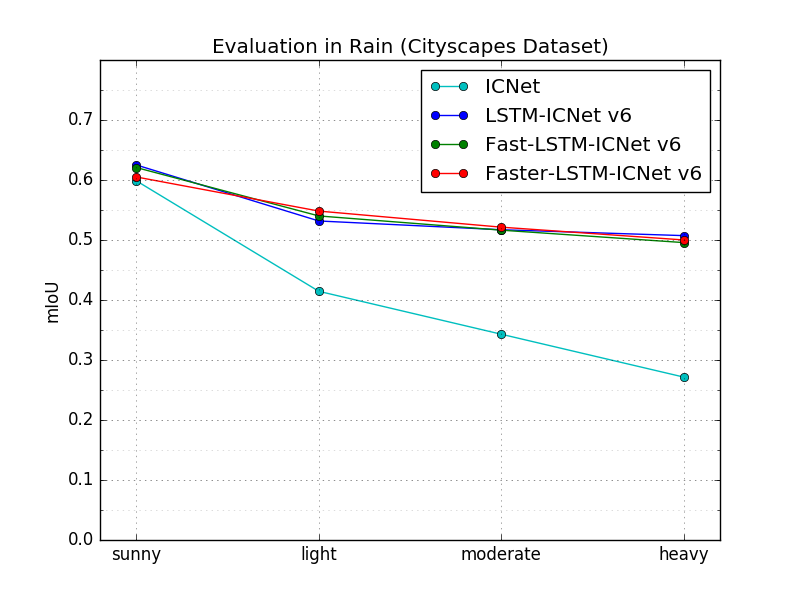}
		\caption{Performance course (mIoU) of the proposed approaches, if the last frame of the video-sequence is disturbed by rain}
		\label{fig_results_cityscape_rain_lastImageDisturbed}
		\vspace{-5mm}
	\end{figure}

\begin{figure*}[tp]
		\includegraphics[width=1.0\textwidth]{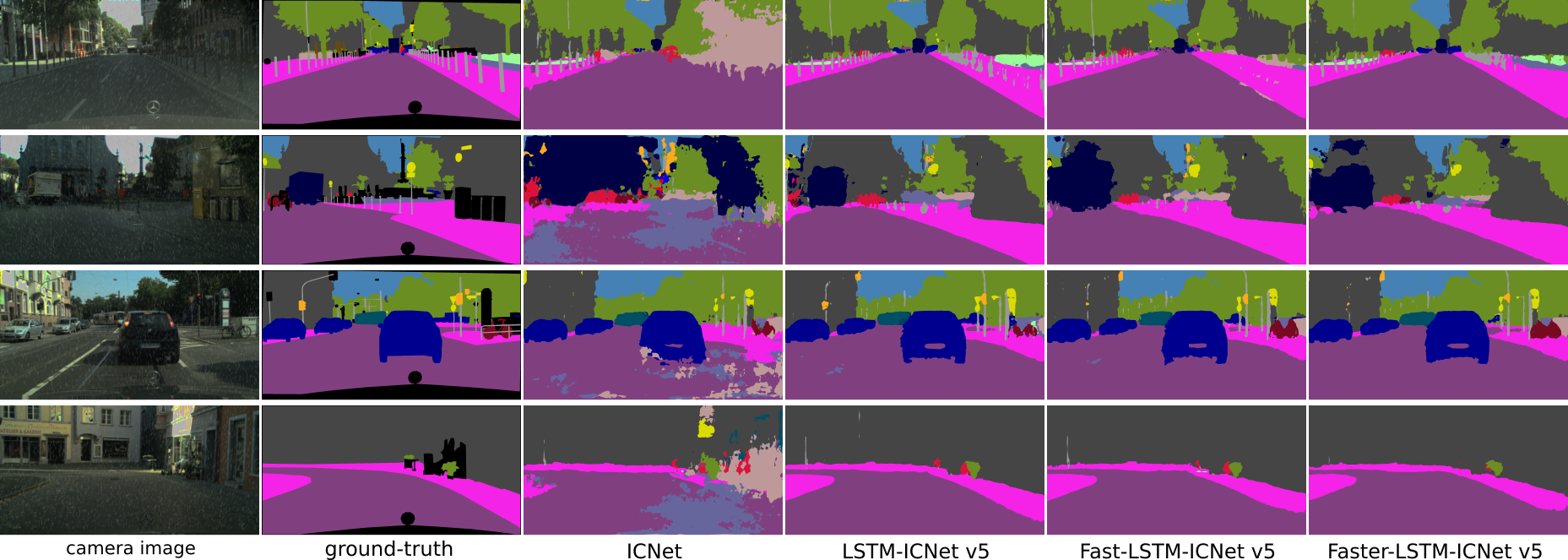}
		\caption{Qualitative results of the proposed approaches on the Cityscapes dataset, if the last frame of the video-sequence is disturbed by heavy rain.}
		\label{fig_Cityscapes_rainLastImage_examples}
	\end{figure*}

	 In the following, the robustness of the proposed video-segmentation approaches is compared with the ICNet, if one frame of the video-sequence is disturbed by unknown noise. More precisely, the last frame of the video-sequences of the Cityscapes dataset is disturbed by adding simulated rain to it, while the remaining images are unmodified. The rain is simulated analogously to \cite{Pfeuffer_2019_RobustSemanticSegmentationInAdverseWeatherConditionsByMeansOfSensorDataFusion}. First, N small lines of pixel length l are randomly drawn in the camera image using a constant slant for one image. Then, the brightness of the image is decreased by 30\%, because rainy days are commonly darker than sunny days. For evaluation, various rain intensities are considered, namely light (N = 500, l = 10), moderate (N = 1500, l = 30), and heavy (N = 2500, l = 60) rain. 
	 In \fig{fig_results_cityscape_rain_lastImageDisturbed}, the mIoU course of the ICNet and the video-segmentation approaches are compared for different rain intensities, and  the results of heavy rain are given in \tab{table_results_Cityscapes}. It turned out that the performance of the ICNet decreases enormously, the heavier the rain is. In contrast, the performance of the remaining approaches decline only slightly. For instance, in the case of heavy rain, the Faster-LSTM-ICNet \mbox{version 6} outperforms the original ICNet by about 23 percent in terms of mIoU and by about 22 percent in terms of pixelwise accuracy, although both approaches achieve similar performance in good weather conditions.  
	 Furthermore, it turns out that the Fast-LSTM-ICNet and Faster-LSTM-ICNet versions perform similarly than the corresponding LSTM-ICNet versions despite its much lower inference time. 
	 In \fig{fig_Cityscapes_rainLastImage_examples}, some qualitative examples of the proposed approaches in heavy rain conditions are given, which show that the ICNet has problems with these disturbances. For example, only some parts of the sidewalk are predicted correctly, and several large ghost objects occur in the images. In contrast, the LSTM-ICNet performs quite good, and classifies most part of the image properly. Moreover, there is hardly a visual difference between the results of the LSTM-ICNet, the Fast-LSTM-ICNet and the Faster-LSTM-ICNet.

	Furthermore, it is of interest, how robust the proposed segmentation approaches are, if every frame of the video-sequence is disturbed by unknown noise, and hence, we add simulated rain to each frame of the video-sequence. \tab{table_results_Cityscapes} contains the corresponding results, which show that the video-segmentation approaches also outperform the basic ICNet. However, their performance also decrease enormously, and their difference to the ICNet is not as large as for the case, in which only one frame of the video-sequence is disturbed.
	All in all, these experiments show that the video-segmentation approaches such as the Faster-LSTM-ICNet are more robust against disturbances, especially, if the disturbances occur in one or a few frames of the video-sequence.

	\begin{table*}[t]
		\caption{Evaluation on the Cityscapes Dataset}
		\begin{center}
			\begin{tabular}{|c||c||c|c||c|c||c|c|}
				\hline
				& & \multicolumn{2}{|c||}{sunny} & \multicolumn{2}{|c||}{heavy rain (last frame)}  & \multicolumn{2}{|c|}{heavy rain (all frames)}  \Tstrut \Bstrut \\ 
				Approach & inference time & acc. (\%) & mIoU (\%) & acc. (\%) & mIoU (\%) & acc. (\%) & mIoU (\%)  \Tstrut \Bstrut \\ \hline \Tstrut
				ICNet \cite{Zhao_2017_ICNet_forRealTimeSemanticSegmentationOnHighResolutionImages}  & $44.66 ms$ & $92.31 \pm 0.08$  & $59.85 \pm 0.47$  & $67.31 \pm 4.51 $ & $27.16 \pm 1.28$ & $67.31 \pm 4.51 $ & $27.16 \pm 1.28$  \Mstrut \\
				\hline \Tstrut
				LSTM-ICNet v2  \cite{Pfeuffer_2019_SemanticSegmentationOfVideoSequencesWithConvolutionalLSTMs} & $60.13 ms$ & $92.96 \pm 0.09$ & $62.00 \pm 0.62$ & $\mathbf{89.94 \pm 0.45}$ & $\mathbf{50.51 \pm 1.51}$ & $\mathbf{79.99 \pm 2.39}$ & $\mathbf{32.95 \pm 2.36}$ \Mstrut \\
				Fast-LSTM-ICNet v2 & $54.64 ms$ & $92.68 \pm 0.28$ & $60.66 \pm 0.79$ & $88.37 \pm 1.09$ & $46.90 \pm 2.09$ & $76.40 \pm 3.36$ & $30.84 \pm 1.82$ \Mstrut \\
				Faster-LSTM-ICNet v2 & $52.11 ms$ & $92.41 \pm 0.14$ & $60.00 \pm 0.81$ & $88.79 \pm 0.61$ & $48.12 \pm 0.74$ & $78.43 \pm 3.02$ & $32.37 \pm 1.41$ \Mstrut \\
				\hline \Tstrut
				LSTM-ICNet v5 \cite{Pfeuffer_2019_SemanticSegmentationOfVideoSequencesWithConvolutionalLSTMs} & $62.14 ms$ & $\mathbf{93.08 \pm 0.07}$ & $62.58 \pm 0.17$ & $88.49 \pm 1.40$ & $48.79 \pm 1.61$ & $66.62 \pm 5.20$ & $28.42 \pm 1.65$ \Mstrut \\
				Fast-LSTM-ICNet v5 & $55.36 ms$ & $92.92 \pm 0.05$ & $62.09 \pm 0.40$ & $88.32 \pm 0.32$ & $48.17 \pm 1.45$ & $69.50 \pm 2.57$ & $29.35 \pm 1.34$ \Mstrut \\
				Faster-LSTM-ICNet v5 & $53.18 ms$ & $92.71 \pm 0.10$ & $60.44 \pm 0.81$ & $88.60 \pm 0.80$ & $49.07 \pm 0.99$ & $70.87 \pm 4.30$ & $29.17 \pm 1.16$ \Mstrut \\
				\hline \Tstrut
				LSTM-ICNet v6 \cite{Pfeuffer_2019_SemanticSegmentationOfVideoSequencesWithConvolutionalLSTMs} & $72.10 ms$ & $93.07 \pm 0.11$ & $\mathbf{62.65 \pm 0.91}$ & $89.07 \pm 1.51$ & $50.40 \pm 1.98$ & $67.87 \pm 5.54$ & $28.03 \pm 1.08$  \Mstrut \\
				Fast-LSTM-ICNet v6 & $63.51 ms$ & $92.93 \pm 0.11$ & $62.07 \pm 1.05$ & $88.71 \pm 1.17$ & $49.57 \pm 1.58$ & $70.63 \pm 6.50$ & $30.16 \pm 1.41$  \Mstrut \\
				Faster-LSTM-ICNet v6 & $55.80 ms$ & $92.73 \pm 0.10$ & $60.71 \pm 0.50$ & $89.15 \pm 0.53$ & $50.28 \pm 1.39$ & $70.15 \pm 6.14$ & $29.18 \pm 1.85$ \Mstrut \\
				\hline
			\end{tabular}
			\label{table_results_Cityscapes}
		\end{center}
	\end{table*}

\subsection{Evaluation in Real Adverse Weather Conditions}

	In the previous section, the proposed segmentation approaches were evaluated on simulated disturbances on the Cityscapes dataset. Now, the proposed methods are evaluated in different real-adverse-weather conditions such as fog, rain or night on the AtUlm-dataset. \tab{table_results_Dense} contains the corresponding results, which show that the video-segmentation approaches also outperform the ICNet in good-weather conditions analogously to the Cityscapes dataset. For example, the LSTM-ICNet \mbox{version 5} and Fast-LSTM-ICNet \mbox{version 5} outperform the ICNet by more than $2$ percent in terms of mIoU, and the Faster-LSTM-ICNet \mbox{version 5} by about $0.8$ percent. 
	Furthermore, the results show that the difference between single-image-segmentation and video-segmentation approaches is much greater in adverse weather conditions. For instance, the performance is increased by about $13\%$ in terms of pixelwise accuracy and by about $9\%$ in terms of mIoU at night, and by about $13\%$ in terms of pixelwise accuracy and by about $5\%$ in terms of mIoU in rain, if the video-segmentation approaches are used instead of the ICNet. Moreover, Fast-LSTM-ICNet and Faster-LSTM-ICNet perform similarly than the corresponding LSTM-ICNet versions in these weather conditions despite its lower inference time.
	
\begin{figure*}[tbp]
		\includegraphics[width=1.0\textwidth]{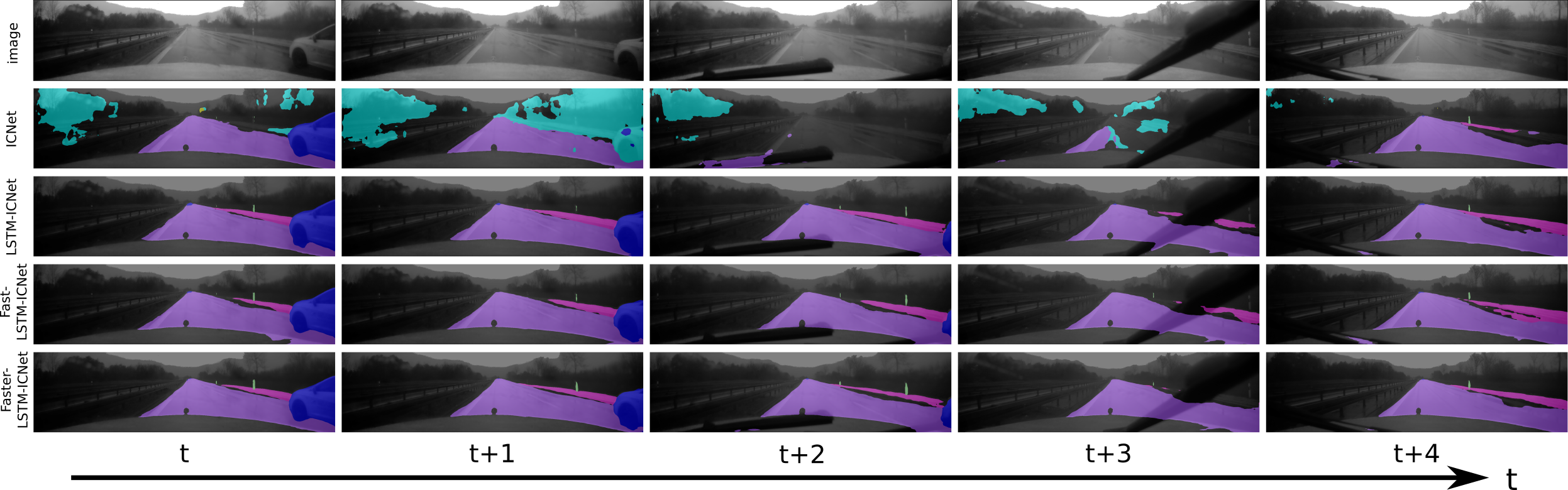}
		\caption{Segmentation maps of a video sequence yielded by the proposed approaches. First row: input image; second row: results of the ICNet; third row: results of the LSTM-ICNet \mbox{version 5}; fourth row: results of the Fast-LSTM-ICNet \mbox{version 5}; fifth row: results of the Faster-LSTM-ICNet \mbox{version 5}}
		\label{fig_example_timeSequence}
	\end{figure*}

	The reason for the performance enhancement is that the segmentation results of the video-segmentation approaches are flickering much less than the results of the ICNet, especially in adverse weather conditions. For instance, the ICNet has problems to detect the road reliably in rain during the entire video sequence (see \fig{fig_example_timeSequence}) and parts of the background are misclassified as truck in some frames. In contrast, LSTM-ICNet and its two modifications perform well in the entire video-sequence, and the neighboring frames do not flicker much.
	This observation is also confirmed by a quantitative analysis, where the amount of flickering pixels of a rainy video sequence is measured by means of the evaluation metric mean Flickering Image Pixels \cite{pfeuffer_2019_SeparableConvolutionalLSTMsForFasterVideoSegmentation} (mFIP). According to \tab{table_results_mFIP}, the result of the ICNet is flickering much more than the proposed video-segmentation approaches, since the mFIP-value of the ICNet is four times higher than the corresponding values of the video-segmentation approaches, while the video-segmentation approaches behave in a similar manner.


\section{Conclusion}
	
	\begingroup
	\setlength{\tabcolsep}{5pt}
	\begin{table*}[t]
		\caption{Evaluation on the AtUlm Dataset}
		\begin{center}
			\begin{tabular}{|c||c|c||c|c||c|c||c|c|}
				\hline
				& \multicolumn{2}{|c||}{sunny} & \multicolumn{2}{|c||}{night}  & \multicolumn{2}{|c||}{rain} & \multicolumn{2}{|c|}{fog} \Tstrut \Bstrut \\ 
				Approach & acc. (\%) & mIoU (\%) & acc. (\%) & mIoU (\%) & acc. (\%) & mIoU (\%) & acc. (\%) & mIoU (\%) \Tstrut \Bstrut \\ \hline \Tstrut
				ICNet \cite{Zhao_2017_ICNet_forRealTimeSemanticSegmentationOnHighResolutionImages}  & $95.17 \pm 0.07$ & $51.61 \pm 0.68$ & $56.81 \pm 1.14$ & $8.00 \pm 0.44$ & $83.34 \pm 1.03$ & $29.35 \pm 2.30$ & $94.63 \pm 0.37$ & $42.15 \pm 1.13$ \Mstrut \\
				\hline \Tstrut
				LSTM-ICNet v2 \cite{Pfeuffer_2019_SemanticSegmentationOfVideoSequencesWithConvolutionalLSTMs} &  $95.11 \pm 0.14$ & $53.08 \pm 2.41$ & $69.65 \pm 1.33$ & $17.81 \pm 1.14$ & $88.06 \pm 0.54$ & $33.48 \pm 2.07$ & $94.99 \pm 0.15$ & $41.92 \pm 1.88$ \Mstrut \\
				Fast-LSTM-ICNet v2 & $95.17 \pm 0.14$ & $52.40 \pm 1.79$ & $68.68 \pm 1.34$ & $16.96 \pm 1.12$ & $88.12 \pm 0.79$ & $32.78 \pm 1.21$ & $95.09 \pm 0.10$ & $41.81 \pm 1.06$ \Mstrut \\
				Faster-LSTM-ICNet v2 & $95.08 \pm 0.11$ & $53.32 \pm 1.07$ & $69.31 \pm 1.23$ & $17.52 \pm 0.82$ & $88.01 \pm 0.41$ & $31.38 \pm 2.87$ & $94.81 \pm 0.14$ & $40.95 \pm 1.15$ \Mstrut \\
				\hline \Tstrut
				LSTM-ICNet v5 \cite{Pfeuffer_2019_SemanticSegmentationOfVideoSequencesWithConvolutionalLSTMs} & $95.24 \pm 0.09$ & $\mathbf{54.81 \pm 0.83}$ & $68.52 \pm 0.90$ & $17.47 \pm 0.61$ & $ 88.35 \pm 0.49$ & $35.58 \pm 1.26$ & $95.29 \pm 0.16$ & $43.41 \pm 1.22$ \Mstrut \\
				Fast-LSTM-ICNet v5 & $95.16 \pm 0.10$ & $54.74 \pm 1.47$ & $69.42 \pm 1.13$ & $17.62 \pm 0.75$ & $88.13 \pm 0.56$ & $33.80 \pm 1.63$ & $\mathbf{95.36 \pm 0.13}$ & $\mathbf{43.74 \pm 1.50}$ \Mstrut \\
				Faster-LSTM-ICNet v5 & $95.10 \pm 0.07$ & $52.47 \pm 1.85$ & $69.27 \pm 1.06$ & $16.63 \pm 0.36$ & $87.83 \pm 0.58$ & $33.84 \pm 2.19$ & $94.93 \pm 0.32$ & $42.83 \pm 0.91$ \Mstrut \\
				\hline \Tstrut
				LSTM-ICNet v6 \cite{Pfeuffer_2019_SemanticSegmentationOfVideoSequencesWithConvolutionalLSTMs} & $\mathbf{95.27 \pm 0.09}$ & $54.33 \pm 1.30$ & $69.96 \pm 1.16$ & $\mathbf{18.12 \pm 1.19}$ & $\mathbf{88.85 \pm 0.51}$ & $\mathbf{36.02 \pm 1.94}$ & $95.30 \pm 0.04$ & $43.33 \pm 0.58$ \Mstrut \\
				Fast-LSTM-ICNet v6 & $95.11 \pm 0.17$ & $54.63 \pm 2.60$ & $68.05 \pm 0.61$ &	$16.72 \pm 0.85$ & $87.98 \pm 1.12$ &	$32.88 \pm 1.58$ &	$95.09 \pm 0.22$ & $42.09 \pm 2.02$ \Mstrut \\
				Faster-LSTM-ICNet v6 & $95.17 \pm 0.14$ & $52.81 \pm 1.87$ & $\mathbf{70.50 \pm 0.63}$ & $17.83 \pm 0.61$ & $88.46 \pm 0.54$ & $34.32 \pm 2.55$ & $95.12 \pm 0.15$ & $42.58 \pm 1.38$ \Mstrut \\
				\hline
			\end{tabular}
			\label{table_results_Dense}
		\end{center}
	\end{table*}
	\endgroup

	\begin{table}[t]
		\caption{Flickering of a rainy video sequence}
		\begin{center}
			\begin{tabular}{|c|c|}
				\hline
				Approach & mFIP (\%)  \Tstrut \Bstrut \\ \hline \Tstrut
				ICNet  & $12.10 \pm 1.23$ \Mstrut \\
				\hline \Tstrut
				LSTM-ICNet v5 & $\mathbf{2.85 \pm 0.45}$  \Mstrut \\
				Fast-LSTM-ICNet v5 & $2.99 \pm 0.28$ \Mstrut \\
				Faster-LSTM-ICNet v5 & $3.07 \pm 0.39$ \Mstrut \\
				\hline
			\end{tabular}
			\label{table_results_mFIP}
		\end{center}
	\end{table}
	
	In this paper, the video-segmentation approach LSTM-ICNet was sped up by two different modifications. It turned out, that the computation time of the LSTM-ICNet can be reduced by up to 23 percent without loss of performance. 	
	Furthermore, the robustness of the proposed approaches were evaluated in simulated and real adverse weather conditions. Evaluation on different datasets and various weather conditions show that the proposed video-segmentation approaches are more robust against disturbances than the single-image segmentation approach, especially, if only a few frames of the video-sequences are disturbed by adverse weather effects. 
	Additionally, the frames of the video-sequence are classified more constantly, and the amount of flickering image pixels is reduced enormously due to the use of temporal image information.

\bibliographystyle{plain}
\bibliography{/home/andreas/Documents/Literatur/Jabref-Datebase/Literatur_Promotion}

\begin{thebibliography}{10}

\bibitem{tensorflow}
Mart\'{\i}n Abadi, Ashish Agarwal, Paul Barham, and Eugene~Brevdo et~al.
\newblock {TensorFlow}: Large-scale machine learning on heterogeneous systems,
  2015.
\newblock Software available from tensorflow.org.

\bibitem{Chen_2018_Deeplabv3p_EncoderDecoderWithAtrousSeparableConvolutionForSemanticImageSegmentation}
Liang-Chieh Chen, Yukun Zhu, George Papandreou, Florian Schroff, and Hartwig
  Adam.
\newblock Encoder-decoder with atrous separable convolution for semantic image
  segmentation.
\newblock In {\em ECCV}, 2018.

\bibitem{cityscape_dataset}
Marius Cordts, Mohamed Omran, Sebastian Ramos, Timo Rehfeld, Markus Enzweiler,
  Rodrigo Benenson, Uwe Franke, Stefan Roth, and Bernt Schiele.
\newblock The cityscapes dataset for semantic urban scene understanding.
\newblock {\em CoRR}, abs/1604.01685, 2016.

\bibitem{pfeuffer_2019_SeparableConvolutionalLSTMsForFasterVideoSegmentation}
A.~{Pfeuffer} and K.~{Dietmayer}.
\newblock Separable convolutional lstms for faster video segmentation.
\newblock In {\em 2019 IEEE Intelligent Transportation Systems Conference
  (ITSC)}, pages 1072--1078, Oct 2019.

\bibitem{Pfeuffer_2019_RobustSemanticSegmentationInAdverseWeatherConditionsByMeansOfSensorDataFusion}
Andreas Pfeuffer and Klaus Dietmayer.
\newblock Robust semantic segmentation in adverse weather conditions by means
  of sensor data fusion.
\newblock In {\em 2019 22nd International Conference on Information Fusion
  (FUSION) (FUSION 2019)}, Ottawa, Canada, July 2019.

\bibitem{Pfeuffer_2019_SemanticSegmentationOfVideoSequencesWithConvolutionalLSTMs}
Andreas Pfeuffer, Karina Schulz, and Klaus Dietmayer.
\newblock Semantic segmentation of video sequences with convolutional lstms.
\newblock In {\em 2019 IEEE Intelligent Vehicles Symposium (IV)}, pages 1253 --
  1259, 2019.

\bibitem{Porav_2019_ICanSeeClearlyNow_ImageRestorationViaDeRaining}
Horia Porav, Tom Bruls, and Paul Newman.
\newblock I can see clearly now : Image restoration via de-raining.
\newblock {\em CoRR}, abs/1901.00893, 2019.

\bibitem{Sakaridis_2018_SemanticFoggySceneUnderstandingWithSyntheticData}
Christos Sakaridis, Dengxin Dai, and Luc Van~Gool.
\newblock Semantic foggy scene understanding with synthetic data.
\newblock {\em International Journal of Computer Vision}, 126(9):973--992, Sep
  2018.

\bibitem{Nabavi_2018_FutureSemanticSegmentationWithConvolutionalLSTM}
Seyed shahabeddin Nabavi, Mrigank Rochan, Yang ~, and Wang ~.
\newblock Future semantic segmentation with convolutional lstm, 07 2018.

\bibitem{Long_2015_FullyConvolutionalNetworksForSemanticSegmentation}
Evan Shelhamer, Jonathan Long, and Trevor Darrell.
\newblock Fully convolutional networks for semantic segmentation.
\newblock {\em CoRR}, abs/1605.06211, 2016.

\bibitem{Shi_2015_ConvolutionalLSTMNetwork_AMachineLearningApproachForPrecipitationNowcasting}
Xingjian Shi, Zhourong Chen, Hao Wang, Dit{-}Yan Yeung, Wai{-}Kin Wong, and
  Wang{-}chun Woo.
\newblock Convolutional {LSTM} network: {A} machine learning approach for
  precipitation nowcasting.
\newblock {\em CoRR}, abs/1506.04214, 2015.

\bibitem{Simonyan_2015_VeryDeepConvolutionalNetworksForLargeScaleImageRecognition}
Karen Simonyan and Andrew Zisserman.
\newblock Very deep convolutional networks for large-scale image recognition.
\newblock {\em CoRR}, abs/1409.1556, 2014.

\bibitem{Valipour_2017_RecurrentFullyConvolutionalNetworksForVideoSegmentation}
Sepehr Valipour, Mennatullah Siam, Martin J{\"a}gersand, and Nilanjan Ray.
\newblock Recurrent fully convolutional networks for video segmentation.
\newblock {\em 2017 IEEE Winter Conference on Applications of Computer Vision
  (WACV)}, pages 29--36, 2017.

\bibitem{Zhao_2017_ICNet_forRealTimeSemanticSegmentationOnHighResolutionImages}
Hengshuang Zhao, Xiaojuan Qi, Xiaoyong Shen, Jianping Shi, and Jiaya Jia.
\newblock Icnet for real-time semantic segmentation on high-resolution images.
\newblock {\em CoRR}, abs/1704.08545, 2017.

\bibitem{Zhao_2017_PyramidScenParsingNetwork}
Hengshuang Zhao, Jianping Shi, Xiaojuan Qi, Xiaogang Wang, and Jiaya Jia.
\newblock Pyramid scene parsing network.
\newblock {\em CoRR}, abs/1612.01105, 2016.

\end{thebibliography}

\end{document}